# Improving Performance of a Group of Classification Algorithms Using Resampling and Feature Selection

Mehdi Naseriparsa
Islamic Azad University, Tehran North Branch
Department Of Computer Engineering
Tehran, Iran

Amir-masoud Bidgoli
Islamic Azad University, Tehran North Branch
MIEEE Manchester University
Tehran, Iran

Touraj Varaee
Islamic Azad University, Tehran North Branch
Tehran, Iran

*Abstract*— In recent years the importance of finding a meaningful pattern from huge datasets has become more challenging. Data miners try to adopt innovative methods to face this problem by applying feature selection methods. In this paper we propose a new hybrid method in which we use a combination of resampling, filtering the sample domain and wrapper subset evaluation method with genetic search to reduce dimensions of Lung-Cancer dataset that we received from UCI Repository of Machine Learning databases. Finally, we apply some well- known classification algorithms (Naïve Bayes, Logistic, Multilayer Perceptron, Best First Decision Tree and JRIP) to the resulting dataset and compare the results and prediction rates before and after the application of our feature selection method on that dataset. The results show a substantial progress in the average performance of five classification algorithms simultaneously and the classification error for these classifiers decreases considerably. The experiments also show that this method outperforms other feature selection methods with a lower cost.

Keywords-Feature Selection; Reliable Features; Lung-Cancer; Classification Algorithms.

## I. INTRODUCTION

Data mining seeks to discover unrecognized associations between data items in an existing database. It is the process of extracting valid, previously unseen or unknown, comprehensible information from large databases. The growth of the size of data and number of existing databases exceeds the ability of humans to analyze this data, which creates both a need and an opportunity to extract knowledge from databases [1].

Assareh[2] proposed a hybrid random model for classification that uses the subspace and domain of the samples to increase the diversity in the classification process. Hayward[3] showed that data preprocessing by choosing suitable features will develop the performance of classification algorithms. In another attempt Duangsoithong and Windeatt [4] presented a method for reducing dimensionality in the datasets which have huge amount of attributes and few samples. Dhiraj[5] used clustering and K-mean algorithm to show the efficiency of this method on huge amount of data. Xiang[6] proposed a hybrid feature selection algorithm that takes the benefit of symmetrical uncertainty and genetic algorithms. Zhou[7] presented a new approach for classification of multi class data. The algorithm performed well on two kind of cancers. Fayyad[8] tried to adopt a method to seek effective features of dataset by applying a fitness function to the attributes. Bidgoli and Naseriparsa[9] proposed a hybrid feature selection method by combination of resampling, chi-squared and consistency evaluation techniques.

Most of the feature selection methods just focus on improving one specific classification algorithm performance. In this paper, we try to improve a group of classification algorithms performance by combining sample domain filtering, resampling and feature subset evaluation methods. We test the performance of the group of classification algorithms on Lung-Cancer dataset.

In section II, III, IV, V, and VI we focus on the definition of feature selection, SMOTE, wrapper method, information gain and genetic algorithm which are used in our proposed method. In section VII, we describe our hybrid method and explain the two phases involved in the feature selection process. In section VIII, the experiments conditions and results are presented. In section IX the proposed method is tested and performance evaluation parameters are calculated. Conclusions are given in section X.

## II. FEATURE SELECTION

Feature selection includes conversion of the main dataset to a new dataset and simultaneously reducing dimensionality by extracting the most suitable features. Conversion and dimensionality reduction will result in a better understanding of the existing patterns in the dataset and more reliable classification by observing the most important data which keeps the maximum properties of the main data. Feature selection consists of four basic steps (Figure 1): subset generation, subset evaluation, stopping criterion, and result validation [10].

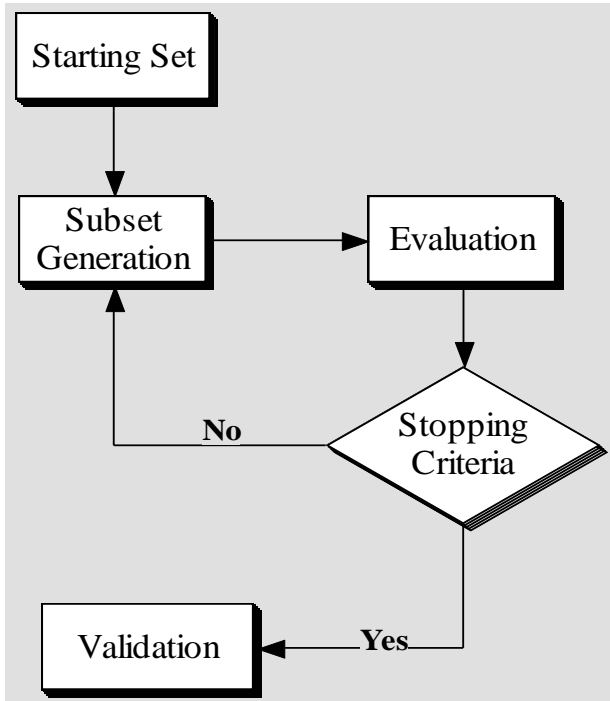

Figure. 1 Feature selection structure

## III. SMOTE: SYNTHETIC MINORITY OVER-SAMPLING TECHNIQUE

Often real world datasets are predominantly composed of normal examples with only a small percentage of abnormal or interesting examples. It is also the case that the cost of misclassifying an abnormal example as a normal example is often much higher than the cost of the reverse error. Under sampling of the majority (normal) class has been proposed as a good means of increasing the sensitivity of a classifier to the minority class. By combination of over-sampling the minority (abnormal) class and under-sampling the majority (normal) class, the classifiers can achieve better performance than only under-sampling the majority class. SMOTE adopts an over-sampling approach in which the minority class is over-sampled by creating synthetic examples rather than by over-sampling with replacement [11].

## IV. WRAPPER METHOD

In the wrapper approach, the feature subset selection is done using the induction algorithm as a black box. The feature subset selection algorithm conducts a search for a good subset using the induction algorithm itself as part of the evaluation function. The accuracy of the induced classifiers is estimated using accuracy estimation techniques [12]. Wrappers are hypothesis driven. They assign some values to weight vectors, and compare the performance of a learning algorithm with different weight vector. In wrapper method, the weights of features are determined by how well the specific feature settings perform in classification learning. The algorithm iteratively adjust feature weights based on its performance.

## V. INFORMATION GAIN

The information gain of a given attribute X with respect to the class attribute Y is the reduction in uncertainty about the value of Y when we know the value of X. The uncertainty about the value of Y is measured by its entropy, H(Y). The uncertainty about the value of Y when we know the value of X is given by the conditional entropy of Y given X, H(Y|X). the formula is shown in equation 1 :

$$IG = H(Y) - H(Y|X) = H(X) - H(X|Y) \quad (1)$$

IG is a symmetrical measure [13]. The information gained about Y after observing X is equal to the information gained about X after observing Y.

## VI. GENETIC ALGORITHM

The genetic algorithm is a method for solving both constrained and unconstrained optimization problems that is based on natural selection, the process that drives biological evolution [14].
The genetic algorithm uses three main types of rules at each step to create the next generation from the current population:
- Selection rules select the individuals, called parents, that contribute to the population at the next generation.
- Crossover rules combine two parents to form children for the next generation.
- Mutation rules apply random changes to individual parents to form children.

## VII. DEFINITION OF THE PROPOSED METHOD

*A. First phase*

In the first step, the SMOTE technique is applied on the original dataset to increase the samples of the minority class. This step contributes to make a more diverse and balanced dataset. In the second step, sample domain filtering method is applied on the resulting dataset to refine the dataset and omit the unreliable samples which are misclassified by the learning algorithm. The learning algorithm for filtering is Naïve Bayes. Naïve Bayes eliminates misclassified samples which are added to the dataset during the resampling process by a low computational cost.
Finally, the original dataset is merged with the secondary dataset. The resulting dataset keeps all the samples of the original dataset and also has some additional samples which contribute to improve accuracy and performance of a group of classifiers.

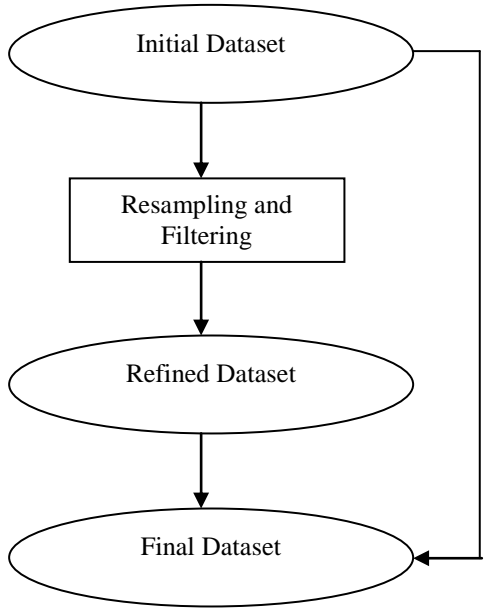

Figure 2 Steps of the first phase

B. *Second Phase*

In the second phase, we focus on the feature space to reach the best subset that results in the best accuracy and performance for the group of classification algorithms. Actually, feature space analysis is carried out in two steps. In the first step, a feature space filtering method is adopted to reduce the feature space and prepare the conditions for the next step. Information gain is a filtering method which uses entropy metric to rank the features and is used for the first filtering step. At the end of this step the features with the ranks higher than the threshold are selected for the next round.

In the second step, wrapper feature selection with genetic search is carried out on the remaining feature subset. Naïve Bayes is chosen as the learning algorithm for wrapper feature selection. The initial population for genetic search is set by the order of features which has been defined by Information gain in the previous step. The features are chosen at the end of this phase are considered as the reliable features.

## VIII. EMPIRICAL STUDY

A. *Experimenatl Conditions*

To evaluate our feature selection method, we choose Lung-Cancer dataset from UCI Repository of Machine Learning databases [15] and apply 5 important classification algorithms before and after implementation of our feature selection method. Lung-Cancer dataset contains 56 features and 32 samples which is classified into three groups. The data described 3 types of pathological lung cancers.

We use WEKA data mining tool to simulate our proposed method and evaluate the performance of classification algorithms. The initial state of the classification algorithms is the default state of WEKA software. In table1, GA parameters are set as follows: Crossover Probability is the probability that two population members will exchange genetic material and is set to 0.6. Max Generations parameter show the number of generations to evaluate and is set to 20. Mutation Probability is the probability of mutation occurring and is set to 0.033 and the last parameter is the number of individuals (attribute sets) in the population that is set to 20.

TABLE I. INITIAL STATES OF GENETIC ALGORITHM

| Parameter | Value |
|---|---|
| Crossover Probability | 0.6 |
| Max Generations | 20 |
| Mutation Probability | 0.033 |
| Population Size | 20 |

B. *Performance Evaluation Parameters Definition*

In table 2, the name and index of the classification algorithms that are used in our experiment are shown.

TABLE II. INDEX AND NAME OF THE CLASSIFICATION ALGORITHMS

| Index | Classification Algorithm Name |
|---|---|
| 1 | Naïve Bayes |
| 2 | Logistic Regression |
| 3 | Multilayer Perceptron |
| 4 | BF Tree |
| 5 | JRIP |

In our experiments, we define some parameters to evaluate our feature selection method. The first parameter, is the number of misclassified samples and we call it MS.

The next parameter, is the average number of misclassified samples of Lung-Cancer dataset on which the classification algorithms applied and we call it AMS. This parameter shows the efficiency of the feature selection method more realistically. AMS parameter formula is shown in equation 2 as:

$$AMS_i = \frac{\sum_{i=1}^{n} MS_i}{N} \qquad (2)$$

In equation 2, $MS_i$, is the number of misclassified samples of Lung-Cancer dataset for a specific classification algorithm. N is the number of classification algorithms which are applied on Lung-Cancer dataset in the experiment.

The third parameter is the relative absolute error of the classification algorithms applied on Lung-Cancer dataset and we show it by RAE.

The next parameter is the average relative absolute error [16] of the classification on Lung-Cancer dataset and we call it ARAE. This parameter shows how a feature selection method could affect the classification algorithms not to predict wrongly or at least their predictions are closer to the correct values. ARAE parameter formula is shown in equation 3 as:

$$ARAE_i = \frac{\sum_{i=1}^{n} RAE_i}{N} \quad (3)$$

In equation 3, $RAE_i$ is the relative absolute error of a specific classification algorithm which is applied on Lung-Cancer dataset. N is the number of classification algorithms which are applied on Lung-Cancer dataset in the experiment.

The next parameters are about correctly and incorrectly classification rates [17] for Lung-Cancer dataset. True positive rate is the rate of correctly classified samples that belong to a specific class in Lung-Cancer dataset and we show it by $TPRate_i$. True negative rate is the rate of correctly classified samples that do not belong to a specific class in Lung-Cancer dataset and we show it by $TNRate_i$. False positive rate is the rate of incorrectly classified samples that do not belong to a specific class in Lung-Cancer dataset and we show it by $FPRate_i$. False negative rate is the rate of incorrectly classified samples that belong to a specific class in Lung-Cancer dataset and we show it by $FNRate_i$. The average true positive, true negative, false positive and false negative are shown in figures 4-7.

$$ATPRate = \frac{\sum_{i=1}^{n} TPRate_i}{N} \quad (4)$$

$$ATNRate = \frac{\sum_{i=1}^{n} TNRate_i}{N} \quad (5)$$

$$AFPRate = \frac{\sum_{i=1}^{n} FPRate_i}{N} \quad (6)$$

$$AFNRate = \frac{\sum_{i=1}^{n} FNRate_i}{N} \quad (7)$$

In equations 4-7, $TPRate_i$, $TNRate_i$, $FPRate_i$ and $FNRate_i$ are true positive, true negative, false positive and false negative for a specific classification algorithm which is applied on the Lung-Cancer dataset respectively. N is the number of Classification algorithms which are applied on Lung-Cancer dataset in the experiment.

## IX. PERFORMANCE EVALUATION

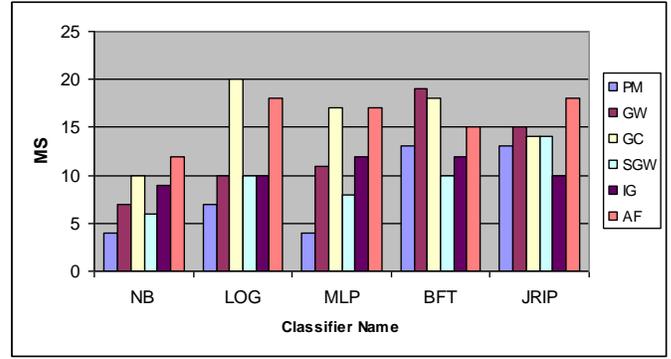

Figure 3 Number of misclassified samples from applying the group of classification algorithms on Lung-Cancer dataset

As it is shown in figure 3, MS parameter is calculated for 5 classification algorithms (Naïve Bayes, Logistic Regression, Multilayer Perceptron, BF Tree, JRIP) and compared with different feature selection methods(GA-Wrapper, GA-Classifier, Symmetrical Uncertainty-GA-Wrapper, Information Gain, All Features). The number of misclassified samples decreases for the group of classifiers when our hybrid feature selection method is applied on Lung-Cancer dataset.

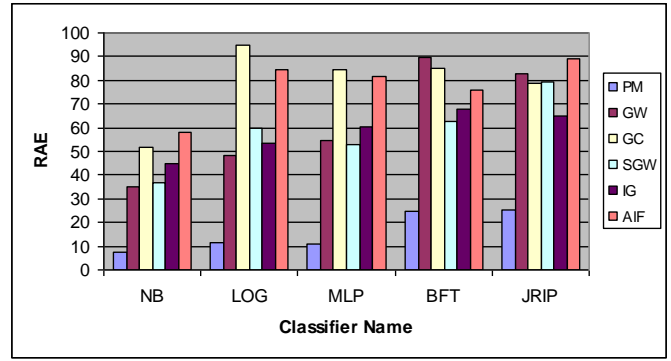

Figure 4 Relative Absolute Error from applying the group of classification algorithms on Lung-Cancer dataset

As it is shown in figure 4, RAE parameter is calculated for 5 classification algorithms and compared with different feature selection methods (GA-Wrapper, GA-Classifier, Symmetrical Uncertainty-GA-Wrapper, Information Gain, All Features). The classification error decreases considerably for the group of classifiers when our hybrid feature selection method is applied on Lung-Cancer dataset.

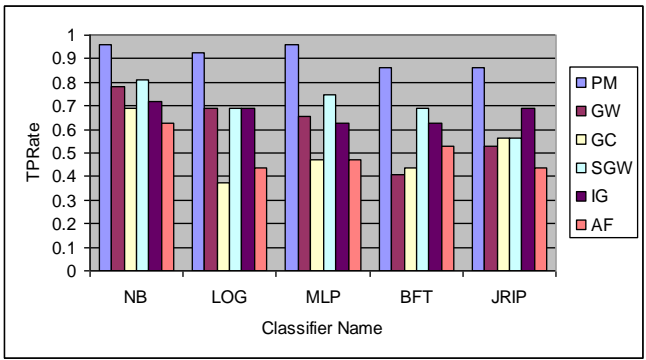

Figure 5 True positive rate from applying the group of classification algorithms on Lung-Cancer dataset

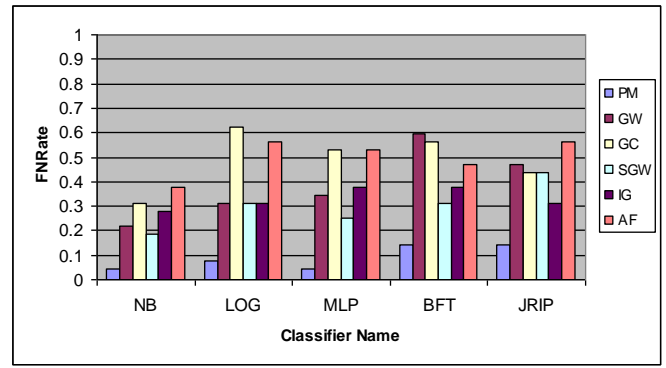

Figure 8 False negative rate from applying the group of classification algorithms on Lung-Cancer dataset

From figure 7 and 8, FPRate and FNRate parameters are calculated for 5 classification algorithms and compared with different feature selection methods (GA-Wrapper, GA-Classifier, Symmetrical Uncertainty-GA-Wrapper, Information Gain, All Features). The rate of incorrectly classified samples for the group of classification algorithms is below 0.2 when our hybrid feature selection method is applied on Lung-Cancer dataset. This shows that our proposed method has decreased the error of the classification process.

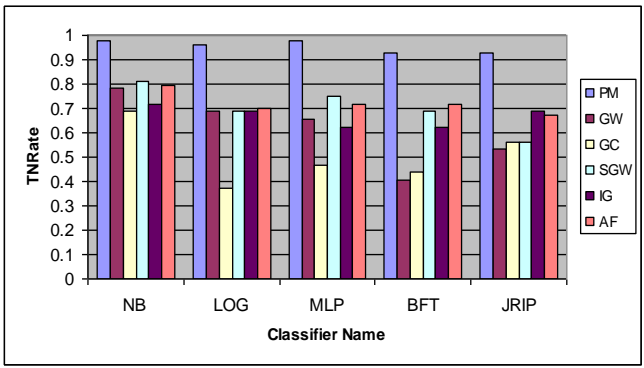

Figure 6 True negative rate from applying the group of classification algorithms on Lung-Cancer dataset

From figure 5 and 6, TPRate and TNRate parameters are calculated for 5 classification algorithms and compared with different feature selection methods (GA-Wrapper, GA-Classifier, Symmetrical Uncertainty-GA-Wrapper, Information Gain, All Features). The rate of correctly classified samples for the group of classification algorithms is above 0.86 when our hybrid feature selection method is applied on Lung-Cancer dataset. This shows that our proposed method has increased the accuracy of the classification process.

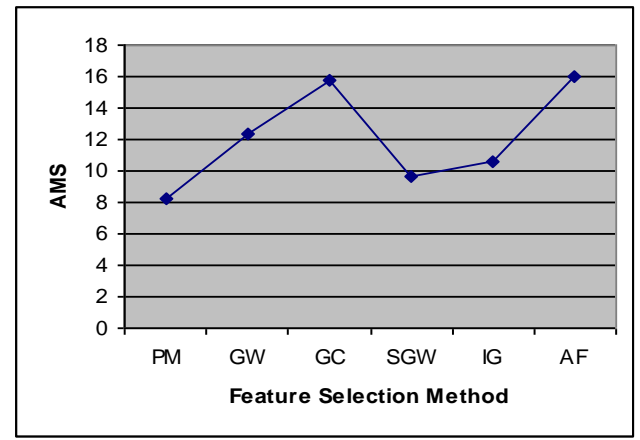

Figure 9 AMS parameter values for different methods

In figure 9, we can see that the AMS parameter for the group of classification algorithms is less than other feature selection methods when our proposed method is applied on Lung-Cancer dataset. This shows that the proposed method is able to improve the accuracy of the group of classifiers simultaneously on Lung-Cancer dataset.

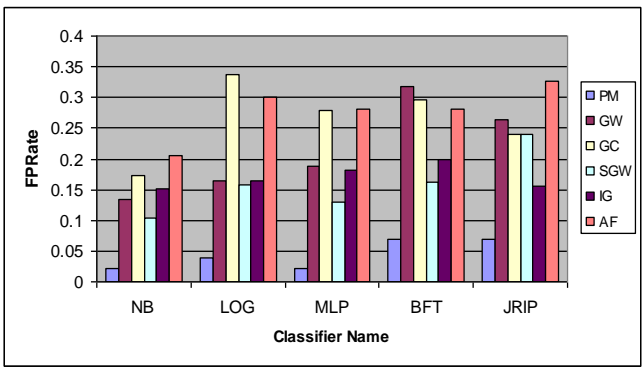

Figure 7 False positive rate from applying the group of classification algorithms on Lung-Cancer dataset

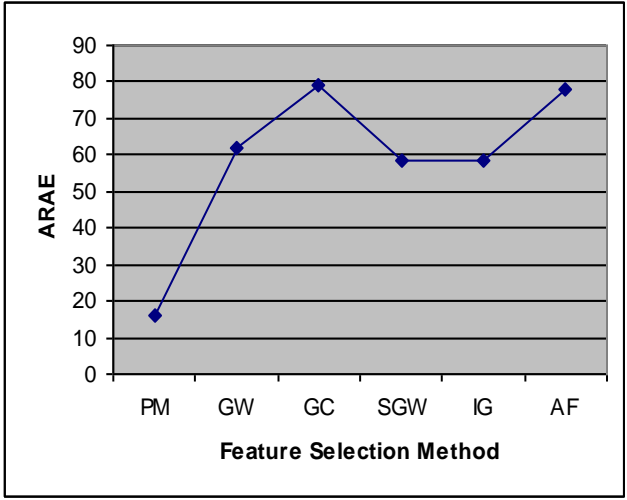

Figure 10 ARAE parameter values for different methods

In figure 10, we can see that the ARAE parameter for the group of classification algorithms is less than 20 percent when our proposed method is applied on Lung-Cancer dataset. This shows that the proposed method is able to decrease the classification error of the group of classification algorithms simultaneously on Lung-Cancer dataset.

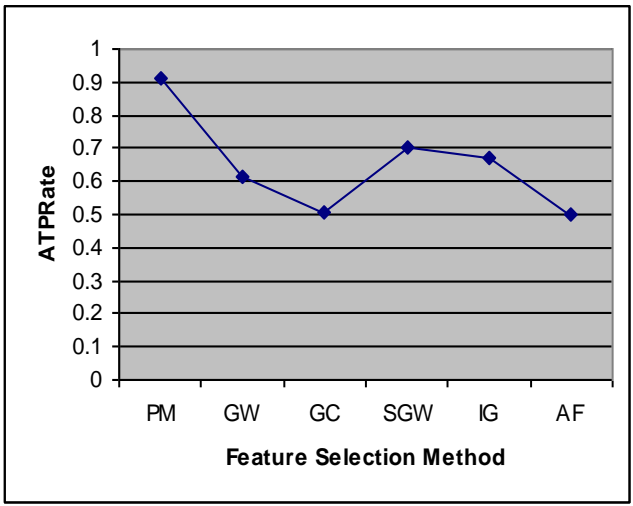

Figure 11 ATPRate parameter values for different methods

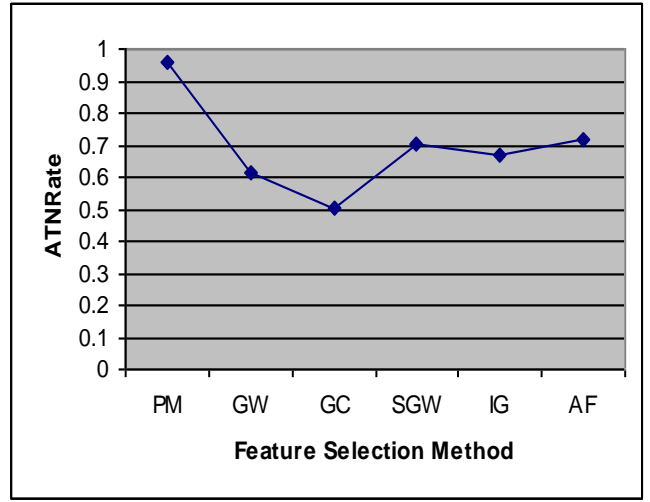

Figure 12 ATNRate parameter values for different methods

In figures 11 and 12, the ATPRate and ATNRate of the group of classification algorithms which are applied on Lung-Cancer dataset are shown. For both figures, the true prediction rate is above 0.9. This shows that the proposed method has increased the true prediction rate of the group of classification algorithms on Lung-Cancer dataset comparing with other methods.

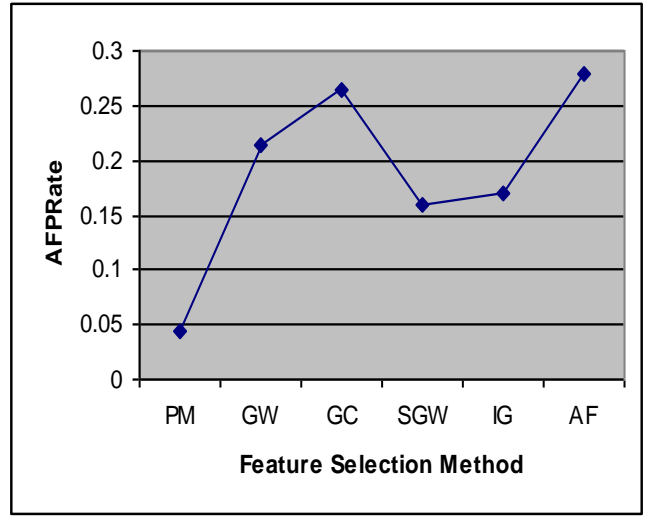

Figure 13 AFPRate parameter values for different methods

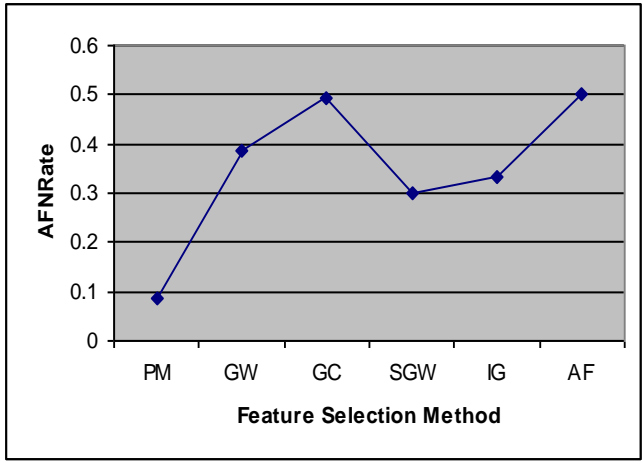

Figure 14 AFNRate parameter values for different methods

In figures 13 and 14, the AFPRate and AFNRate of the group of classification algorithms which are applied on Lung-Cancer dataset are shown. For both figures, the false prediction rate is below 0.1. This shows that the proposed method had decreased the false prediction rate of the group of classification algorithm on Lung-Cancer dataset comparing with other methods.

From the figures above, we observe that the proposed method achieves higher classification accuracy for the group of classification algorithms in comparison to other methods. Moreover, the cost of our proposed method is considerably smaller than the GA-Wrapper and GA-Classifier methods, because it achieves higher level of dimensionality reduction.

## X. CONCLUSION

A hybrid feature selection method is proposed. This method combines resampling and sample filtering with feature space filtering and wrapper methods. The first phase analyses sample domain and in the second phase, feature space filtering eliminates irrelevant features and then wrapper method select reliable features with a lower cost and higher accuracy. Different performance evaluation parameters are defined and calculated on Lung-Cancer dataset. The results show that our proposed method outperforms other feature selection methods (GA-Wrapper, GA-Classifier, Symmetrical Uncertainty-GA-Wrapper, Information Gain, All Features) on Lung-Cancer dataset. Furthermore, the proposed method improves the accuracy and true prediction rate of the group of classification algorithms simultaneously.